# A Scalable and Generalized Deep Learning Framework for Anomaly Detection in Surveillance Videos

Sabah Abdulazeez Jebur [1], Laith Alzubaidi [2,3,*], Khalid A. Hussein [4], Haider Kadhim Hoomod [4], Ahmed Ali Saihood [5], YuanTong Gu [2]

1. Department of Computer Sciences, University of Technology, Baghdad, Iraq.
   sabah.abdulazeez@iku.edu.iq
2. School of Mechanical, Medical, and Process Engineering, Queensland University of Technology Brisbane, QLD 4000, Australia.
   l.alzubaidi@qut.edu.au (L.A); yuantong.gu@qut.edu.au (Y.T.G)
3. Centre for Data Science, Queensland University of Technology Brisbane, QLD 4000, Australia.
4. Department of Computer Science, College of Education, Mustansiriyah University, Baghdad, Iraq.
   dr.khalid.ali68@gmail.com (K.A.H); drhjnew@gmail.com (H.K.H)
5. Faculty of Computer Science and Mathematics, University of Thi-Qar, Nasiriyah 00964, Thi-Qar, Iraq.
   ahmed.alisiehood@gmail.com

*Corresponding author: l.alzubaidi@qut.edu.au

## Abstract

Anomaly detection in videos is challenging due to the complexity, noise, and diverse nature of activities such as violence, shoplifting, and vandalism. While deep learning (DL) has shown excellent performance in this area, existing approaches have struggled to apply DL models across different anomaly tasks without extensive retraining. This repeated retraining is time-consuming, computationally intensive, and unfair. To address this limitation, a new DL framework is introduced in this study, consisting of three key components: transfer learning to enhance feature generalization, model fusion to improve feature representation, and multi-task classification to generalize the classifier across multiple tasks without training from scratch when new task is introduced. The framework's main advantage is its ability to generalize without requiring retraining from scratch for each new task. Empirical evaluations demonstrate the framework's effectiveness, achieving an accuracy of 97.99% on the RLVS dataset (violence detection), 83.59% on the UCF dataset (shoplifting detection), and 88.37% across both datasets using a single classifier without retraining. Additionally, when tested on an unseen dataset, the framework achieved an accuracy of 87.25%. The study also utilizes two explainability tools to identify potential biases, ensuring robustness and fairness. This research represents the first successful resolution of the generalization issue in anomaly detection, marking a significant advancement in the field.

## 1. Introduction

Integrating Artificial Intelligence (AI) in various domains has brought transformative changes in our daily lives. One of the major uses of AI lies in the field of surveillance cameras, which enhance security and situational awareness. Anomaly detection (AD), a cutting-edge AI technique, has emerged as a vital tool in the surveillance industry, revolutionizing how we monitor and protect our environments [1]. The process of AI-driven AD employs machine learning (ML) and deep learning (DL) algorithms to automatically identify unusual behaviors in video streams. These AI systems continuously analyze and learn from massive datasets, allowing them to detect subtle irregularities that might go unnoticed by human observers. This makes them extremely valuable for early threat detection, crime prevention, and overall safety enhancement [2]. AD identifies human activities that deviate from expected behavior, including activities such as violence, stealing, arson, abuse, loitering, and vandalism in specific locations such as markets and streets. Regarding Computer Vision (CV), AD involves recognizing patterns that exhibit significant deviations from normal behavior. Monitoring surveillance cameras without leveraging intelligent systems demands substantial resources, including manpower, finances, and time. Moreover, it is susceptible to errors due to the challenge of simultaneously monitoring numerous surveillance cameras [3]. Automated CV systems are critical for detecting anomalies in video without manual intervention. DL techniques have consistently delivered state-of-the-art results in addressing issues related to monitoring suspicious activities in surveillance systems and excelling in various other domains. DL leverages deep neural networks (DNNs) to identify and extract features from input data. Furthermore, it has the capability to automatically discern numerous unidentified parameters during the training phase [3]. Convolutional Neural Networks (CNNs) are DNNs specifically designed to automatically learn features and recognize patterns from image data. CNNs use a specialized convolutional layer to analyze the input image with small filters or kernels, allowing them to detect different features at multiple scales[4]. These networks have succeeded highly in various image-related tasks, including image recognition and generation [5]. CNNs are very efficient and powerful tools in video anomaly detection (VAD). These DL algorithms can classify video frames based on their feature content and extract relevant information. A critical application of CNNs is identifying abnormal events within video streams. This involves training a CNN on a labeled video dataset to enable it to detect deviations from expected behavior and flag anomalies. The synergy between CNNs and VAD enhances security and showcases the remarkable adaptability of CNNs in extracting meaningful insights from dynamic visual data [6]. However, one challenge that CNNs face is the requirement for a large amount of data. To address this problem, researchers have widely used transfer learning (TL). TL is a technique in which the knowledge gained from training a model on one task is utilized to improve the performance of another related task instead of training a model from scratch for each specific task. In simpler terms, TL uses the knowledge acquired from one task to enhance learning and performance on another [7]. In addition to the data scarcity problem, VAD faces significant challenges related to generalization [5], which is the ability of a model to perform well on new data. In addition, the flexible fusion of multiple models into a single framework also presents a challenge in this field. The fusion approach offers a concise representation of various features extracted from different sources, enhancing overall performance and improving generalization capability [8][9]. In the DL field, integrating new sets of data without requiring extensive retraining is a significant challenge. Furthermore, understanding and interpreting the decisions made by DNNs in the context of image classification and object detection is crucial. Moreover, DNNs, particularly deep convolutional models, are highly complex and can be considered black boxes, making it difficult to understand how they arrive at their predictions[10]. This need for more transparency is a significant concern, especially in critical applications where trust and explainability are essential. Multi-anomaly detection is the task of identifying multiple types or classes of anomalies in a dataset. Unlike binary anomaly detection, which deals with only one class of anomalies and one class of normal data, multi-anomaly detection addresses scenarios with multiple types of anomalies. Incorporating new classes into the existing model typically involves retraining the model from scratch, which can be time-consuming, resource-intensive, and may require a substantial amount of labeled data for the new class. Moreover, ensuring the model can detect all known anomaly types is crucial while adapting to new ones. We can create more efficient and practical anomaly detection systems if we avoid this process. In this study, new methods have been proposed to address the issues mentioned above. As a result, we present the following contributions:

- A novel framework has been introduced for incorporating a new anomaly class into an existing AD model without retraining from scratch.
- A deep feature fusion method is proposed to integrate diverse DL models for better feature representation.
- The deep feature fusion approach achieved an accuracy of 83.59% on the UCF-Crime dataset and 97.99% on the RLVS dataset, outperforming all the previous methods.
- Based on our experiments with the UCF-Crime and RLVS datasets, the multi-classification approach we used was able to accurately detect and categorize two specific abnormal behavior classes (shoplifting and violence) and normal behavior, achieving an accuracy of 88.37%. This approach significantly enhances the ability to identify and categorize abnormal behaviors in different scenarios. To the best of our knowledge, an existing approach has yet to achieve a similar level of performance with a single model across multiple tasks in VAD.
- The proposed framework achieved an accuracy of 87.25% on a complete independent test set.

## 2. Related Works

This section provides an overview of the evolving landscape of VAD research, highlighting key studies and methodologies that harness the potential of ML and DL to enhance the accuracy and efficiency of AD systems. A study [11] presented a new DL architecture for identifying violent behaviors in videos. The method leverages recurrent neural networks (RNNs) and 2D CNN to capture spatial and temporal features. Optical flow information is integrated to encode motion patterns. The proposed approach has been tested with success on multiple databases. In [12], the authors proposed an approach to assist monitoring staff in directing their attention to specific screens where the likelihood of a crime is higher. This approach involved identifying situations in video footage that may signal an impending crime. They employed a 3D CNN to examine surveillance videos and capture behavioral attributes for the identification of suspicious actions. The model was trained using carefully chosen videos from the UCF-Crimes dataset. In [13], a Hybrid CNN Framework (HCF) was presented to identify distracted driver behaviors by leveraging DL and image processing techniques. The framework employed pre-trained CNN models in collaboration to extract behavioral features through TL, thereby improving result accuracy. In a study [14], TL was employed to enhance the accuracy of abnormal behavior detection by extracting human motion characteristics from RGB video frames. The authors utilized the VGGNet-19 architecture for feature extraction and subsequently applied a Support Vector Machine (SVM) classifier to identify complex motion scenarios. In [15], several DL models, such as CNN, LSTM (Long Short-Term Memory), CNN-LSTM, and Autoencoder-CNN-LSTM, were explored to identify unusual behaviors in the elderly. The models were trained using temporal and spatial data, which enabled them to make accurate predictions. To tackle the issue of data imbalance, the researchers oversampled minority classes, especially for the LSTM model. Overall, the study provides insights into how DL can be used to detect and prevent unusual behaviors in elderly individuals. In[16], the authors tackled the issue of shoplifting by focusing on detecting suspicious behaviors that may lead to criminal activities. Instead of identifying the crime itself, their approach aims to model and detect behaviors that precede criminal acts, providing opportunities for prevention. They utilized a 3D CNN to extract video features and classify segments containing potential shoplifting behavior. [17] introduced a shoplifting detection system that utilizes a DNN. The system used the Inceptionv3 model for feature extraction and employed LSTM networks to understand temporal sequences. This system can accurately identify individuals involved in shoplifting activities with an accuracy of up to 74.53%. The paper [18] presented deep violence detection approach that leverages handcrafted features related to appearance, speed of movement, and representative images. These features are input into CNN through spatial, temporal, and spatiotemporal streams. The spatial stream captured environment patterns, the temporal stream focused on motion patterns using modified optical flow, and the spatiotemporal stream introduced a novel feature to enhance interpretability. The CNN is trained on datasets containing both violent and normal behavior frames. A study [19] used DL techniques to detect abnormal driver behaviors such as smoking, eating, drinking, and calling. To train and test models, a dataset was created comprising these behaviors as well as normal driving. The study evaluated DL models, including a proposed CNN-based model and pre-trained models such as ResNet101, VGG-16, VGG-19, and Inception-v3. Keyframe extraction was used to optimize computation. In [20], a shoplifting detection system was introduced. This approach involved the use of a hybrid neural network that combined convolutional and recurrent components to extract information from video frames and analyze their temporal sequence. Specifically, it employed gated recurrent units for data processing. Data augmentation was conducted to mitigate class imbalance and enhance the dataset. For classification, a pre-trained MobileNetV3Large CNN was combined with a recurrent network that incorporated gated nodes. In reference [21], a method for detecting violent behavior using keyframes is proposed. This approach treats video frames as discrete events and detects instances of violence by assessing whether the count of keyframes exceeds a predefined threshold, thereby reducing hardware demands. Furthermore, the paper introduced a novel training technique that leverages pairs of background-removed and original images to improve feature extraction for DL models, all while avoiding the introduction of extra network complexity. In [22], a model for detecting crowd violence behavior, named HD-Net, was developed. It utilized a human contour extractor to minimize background noise in violence detection by focusing on individuals in video frames. A dynamic feature encoder is also used for extracting dynamic features from adjacent frames. The model is built on a 3D CNN framework for spatial feature extraction and LSTM for temporal feature fusion. [23] proposed a convolutional autoencoder architecture that can detect anomalies in appearance and motion patterns. The architecture used two components, the spatial and temporal autoencoder, to differentiate between spatial and temporal representations. The spatial autoencoder captures appearance features by reconstructing the initial frame. On the other hand, the temporal component models motion through RGB differences across sequential frames. To further enhance the performance of the motion autoencoder, the paper incorporated a variance-based attention module that highlights critical movement areas. A novel deep K-means clustering approach was introduced to extract concise representations. In [24], the authors presented a CNN model to detect crowd anomalies in video sequences. The model is composed of two convolutional layers followed by two fully connected layers that utilize Rectified Linear Unit (ReLU) and sigmoid functions. The intermediate layers generate features that are used for abnormality detection. The model's performance was evaluated on three scientific datasets that included normal and abnormal activities. The outcomes demonstrated that the model performed effectively when applied to random YouTube videos exhibiting abnormal behavior. In [25], a new approach was introduced to detect abnormal behavior of workers in manufacturing environments. The model identifies and describes unusual worker actions based on their interaction with objects using a combination of technologies such as Mask R-CNN, Media Pipe Holistic, LSTM, and a worker behavior description algorithm. The approach involved object recognition, worker pose identification, and pattern analysis to differentiate between typical and unusual actions. Anomalous behaviors encompass instances such as worker falls, slips, tool breakage, and machine failures. The article [26] presented an anomaly recognition model employing a deep CNN architecture. This model extracts deep features from surveillance video frames and directs them to a temporal convolution network (TCN) with a multi-head attention module. The TCN comprises multiple layers of temporal convolutional filters with varying dilation rates, enabling the capture of diverse temporal contexts and long-range dependencies. It is trained by minimizing an objective function, using cross-entropy loss, to optimize parameters for accurate classification or recognition of activities in sequential data. The research presented in the paper [27] utilized TL-InceptionV3 to improve anomaly detection in surveillance cameras. Two TL methods, pre-training and fine-tuning, were employed using InceptionV3 to classify frames as normal or abnormal behaviors. The UCF-Crime dataset was utilized for training and evaluation. The results demonstrated that the fine-tuning approach outperformed the pre-training approach significantly. This indicates substantial enhancements in the model's performance. In [28], a comprehensive benchmark dataset was introduced, consisting of 900 samples, evenly divided into 450 instances of shoplifting and 450 instances of non-shoplifting, annotated across different shoplifting scenarios. This dataset was utilized to assess shoplifting detection techniques, including 2D CNN, 3D CNN, and a novel hybrid method that combines InceptionV3 and bidirectional LSTM. Notably, the hybrid approach outperformed the others in terms of performance. In [29], a DL method is introduced for identifying violence in animation videos. The research involved modifying a Faster R-CNN model to handle the intricate aspects of violence depicted in cartoon and animation content. The modification included replacing the model's backbone with a customized RegNet to capture frame features, utilizing a modulated deformable convolutional (MDC) layer instead of the standard inner lateral connection for flexible feature map extraction, and introducing a novel distributed attention module (DAM) within the feature pyramid network to enhance feature extraction. Additionally, to enhance violence detection across diverse scenarios, the researchers incorporated a modified multiscale Region of Interest (ROI) Align. Moreover, the method integrated a classification component into the detection model to categorize different levels of violence within each frame. In [30], an innovative semi-supervised hard attention mechanism was introduced. This mechanism facilitated the identification and separation of crucial regions within videos from less informative segments of the data. By efficiently eliminating redundant data and highlighting valuable visual information at a higher resolution, the model's accuracy saw improvement. This approach obviated the necessity for attention annotations in video violence datasets, rendering them more widely applicable. The proposed model utilized a pre-trained I3D backbone to expedite and stabilize the training process. The ref.[31] proposed a novel approach to enhance the generalization of violence detection across multiple scenarios. This approach employed three pre-trained CNN models, Xception, InceptionV3, and InceptionResNetV2, to extract significant features from RLVS and Hockey datasets. The extracted features from each dataset were then fused into a single feature pool separately. Finally, these feature pools from the first violence scenario and the second were combined into a unified feature space, facilitating the training of an ML classifier capable of generalizing across multiple scenarios. However, it is essential to note that this approach needs to address generalization across multi-task anomaly detection.

The article [32] explored advanced techniques to enhance aggression detection in surveillance systems by utilizing multimodal fusion and DL. The study addressed the limitations of traditional single-modality approaches by integrating audio, visual, and text-based features, along with additional meta-information such as Audio-Focus, Video-Focus, Context, History, and Semantics. Four distinct fusion methods were developed and compared: intermediate level fusion, concatenation-based fusion, and two methods involving element-wise operations followed by concatenation.

In the paper [33], the authors proposed a method to enhance the detection of abnormal actions, particularly human aggression and car accidents, using wavelet-based channel augmentation. The core of the proposed method was the MultiWave-Net, a spatiotemporal network designed to integrate wavelet transformation with traditional DL architectures such as CNNs and ConvLSTMs. The wavelet-based channel augmentation technique was applied to improve the feature extraction capabilities of these networks, allowing them to better capture both spatial and temporal aspects of the input data.

The article [34] proposed an approach to recognizing violent behaviors in videos by leveraging the MLP-Mixer architecture and a new dataset format called Sequential Image Collage (SIC). SIC aggregated video frames into sequential image collages, capturing both spatial and temporal dimensions to enhance the model's understanding of violent actions. These collages, along with original frames, were processed through the MLP-Mixer architecture, which relied solely on Multilayer Perceptrons (MLPs) for computational efficiency. The method involved patch embedding, token mixing, and channel mixing operations to capture both local and global features from the dataset.

Despite the literature introducing advanced AI methods to detect and recognize anomaly behaviors in videos, a significant challenge emerges concerning generalization. These methods necessitate retraining the entire model from scratch when a new anomaly class is introduced, resulting in increased time and computational resource requirements, especially in cases where the model is complex or requires extensive training. This challenge presents a practicality and efficiency issue for anomaly detection systems. Therefore, new approaches are urgently needed to address the generalization problem in multi-task anomaly detection without requiring extensive retraining.

## 3. Materials and Methods

### 3.1 Datasets

UCF-Crime dataset and RLVS dataset were employed in this study. The UCF dataset [35] is widely recognized in criminal activity recognition research and was compiled by the University of Central Florida (UCF). It is publicly accessible for research purposes and includes 1,900 untrimmed surveillance videos gathered from platforms such as YouTube, TV news, and documentaries, varying in quality and resolution. This dataset encompasses 128 hours of footage, depicting 13 real-world criminal activities, including abuse, arrest, arson, assault, road accidents, shoplifting, and more. For this investigation, samples were specifically extracted from the "shoplifting" and "normal" classes, each comprising 50 video clips recorded in retail stores. On the other hand, the RLVS dataset [36] consists of 2,000 video clips, equally divided between violent and normal activities. The violent videos depict physical altercations in various environments, such as streets, prisons, and schools. Videos within the RLVS dataset feature high resolutions, ranging from 480p to 720p, with durations spanning 3 to 7 seconds. A frame extraction process was performed using a 10-frame interval, resulting in six frames per second. Notably, some frames within violent and shoplifting videos were eliminated during the data cleaning phase because they did not depict the relevant actions and were more similar to frames from normal videos. Figure 1 shows some samples from both the UCF and RLVS datasets.

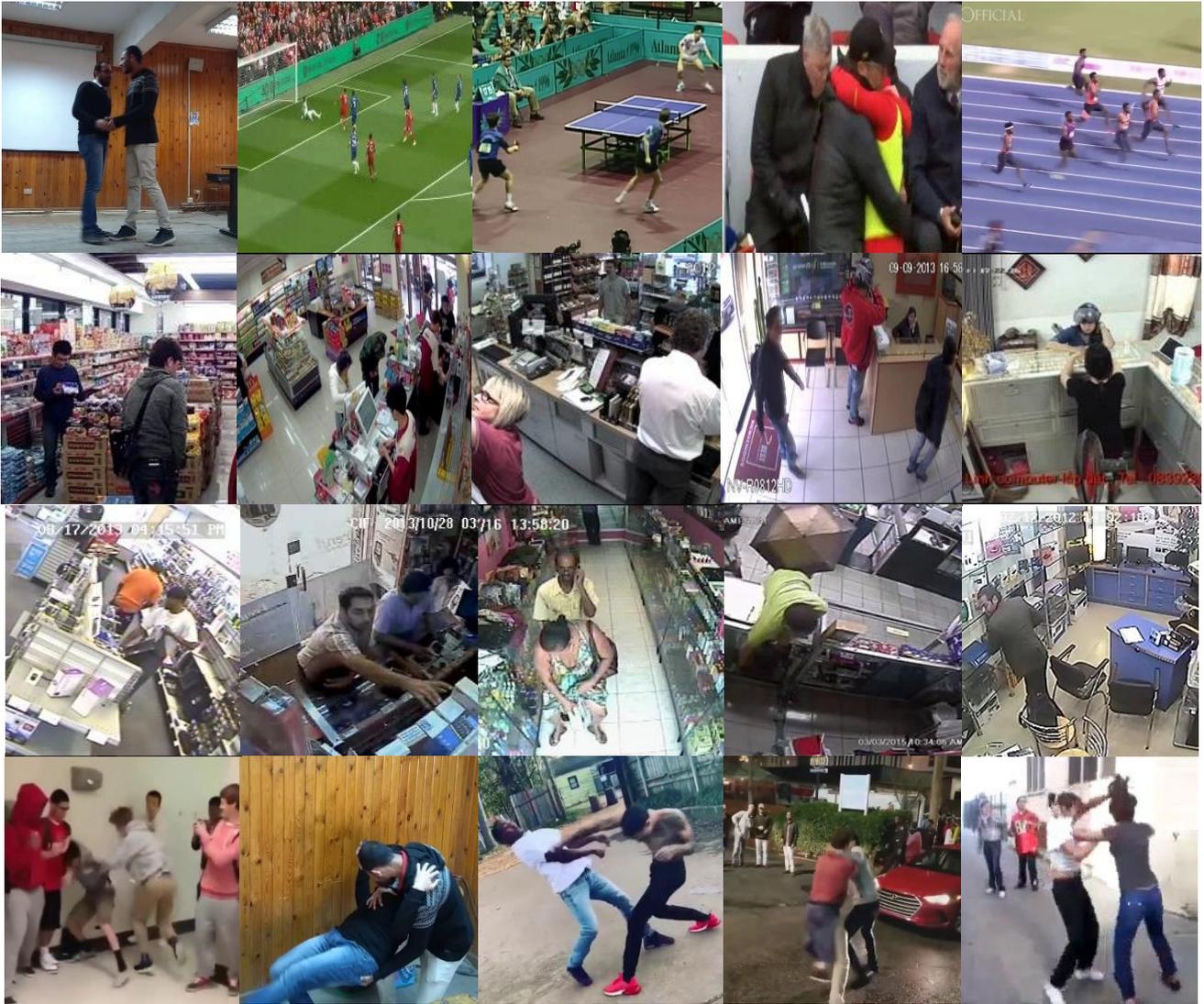

Figure 1. Samples from the UCF and RLVS datasets. The first two rows depict normal samples, while the last two depict shoplifting and violence samples in both datasets.

## 3.2 CNN Architectures

This study utilized four deep CNN models, MobileNetV2, InceptionV3, InceptionResNetV2, and Xception, to tackle the challenge of detecting anomalous video behaviours. These models have several advantages worth noting. They have shown outstanding performance on the ImageNet dataset, a widely recognized benchmark for computer vision tasks. Furthermore, their well-designed architecture excels at feature extraction, allowing them to capture a broad spectrum of features due to their diverse filter sizes, ranging from 1×1 to 7×7. Furthermore, incorporating ReLU activations and residual connections improves the quality of feature representation and addresses the problem of gradient vanishing. Dropout layers and global average pooling (GAP) are also utilized in these models to mitigate the risk of overfitting. Moreover, the incorporation of Batch Normalization layers expedites the training process. These advantages collectively make these models effective methods for VAD. The following subsections provide concise descriptions of the models used in the work.

### 3.2.1 MobileNetV2 model

The MobileNetV2 model is an efficient and lightweight CNN technique explicitly designed for mobile and embedded devices. MobileNetV2's architectural composition commences with fully convolutional layers containing 32 filters and encompasses 19 residual bottleneck layers. It consists of two modules, each comprising three layers. These blocks start and end with a 1 × 1 convolutional layer comprising 32 filters—notably, the second block functions as a fully connected layer with one depth. ReLU activation is applied at various levels throughout the architecture. The primary distinction between the two modules is their stride lengths; the first employs a stride length of 1, while the second utilizes a stride length of 2 [37]. The MobileNetV2 model successfully achieves a delicate equilibrium between model size and performance, rendering it particularly suitable for resource-constrained applications like mobile devices and embedded systems.

### 3.2.2 InceptionV3 Model

InceptionV3 [38] is a CNN architecture for image classification and object recognition tasks. It is renowned for its deep structure and utilization of specialized convolutional layers, including Inception modules designed to capture features at varying scales. InceptionV3 employs parallel convolutional layers of different sizes and pooling operations within these modules to effectively capture features at different scales. It leverages factorized convolutions to reduce network parameters, includes auxiliary classifiers to enhance training, and utilizes GAP to mitigate overfitting. Extensive use of batch normalization accelerates training. InceptionV3's depth and parameter count make it suitable for various CV applications. These attributes contribute to InceptionV3's ability to achieve high accuracy in image classification tasks while maintaining manageable computational complexity.

### 3.2.3 InceptionResNetV2 Model

InceptionResNetV2 is a deep CNN architecture combining elements from both the Inception and ResNet architecture [39]. It was designed to improve feature learning and representation in CV tasks. InceptionResNetV2 integrates residual connections, similar to ResNet, to facilitate the training of deep networks while also utilizing Inception modules to capture features at multiple scales. This architecture typically includes a stem module to preprocess input data, multiple Inception-ResNet blocks that increase network depth, and final layers for classification or feature extraction. InceptionResNetV2's unique combination of these architectural elements aims to achieve superior performance in tasks such as image classification and object detection.

### 3.3.4 Xception Model

The Xception network [40] represents an evolution from Inception by replacing conventional convolution layers with depthwise separable convolution layers. This design optimizes spatial and cross-channel correlations within the network's core functionality. XceptionNet, with 36 convolution layers segmented into 14 modules, supersedes Inception's architecture. It maintains a continuous relationship between the remaining layers after removing the initial and final ones. The network transforms the original image to determine probabilities across multiple input channels and employs 11 depth wise convolutions, offering an alternative to three-dimensional maps by visualizing relationship patterns.

## 3.3 Part1: Proposed framework, Deep Feature Fusion Approach:

In order to leverage the features extracted by different CNN models for detecting anomalies in surveillance videos, since each model has its architecture and different filter sizes for feature extraction from input data, combining these features provides a better feature representation. This, in turn, improves overall performance. The proposed deep feature fusion approach (Figure 2) used four CNN models, MobileNetV2, InceptionV3, InceptionResNetV2, and Xception, as feature extractors to capture features from the input video frames. Next, these features, extracted from the individual models, are combined into a unified feature pool. The different colors in the feature pool correspond to the features extracted from different CNN models. The feature pool is then used to train ML classifiers. Finally, ML classifiers were employed to assign class labels and classify human behaviors as normal or abnormal. Six classifiers were used to recognize anomalies in captured videos: SVM, SoftMax, K-Nearest Neighbor (KNN), AdaBoost, Logistic Regression (LogReg), and Naive Bayes classifiers.

The deep fusion approach offers several benefits. It provides a flexible means of combining multiple CNN models without the need to train them from scratch. This approach allows for the incorporation of new models trained on specific datasets by extracting features from the final fully connected layer and then inserting them into the feature space. This incorporation method saves time and computational costs, eliminating the need to retrain the already-used individual models. Moreover, the fusion of features extracted from multiple models yields a broader and more inclusive array of information for classifiers to acquire knowledge from. This approach empowers ML classifiers to harness the distinct strengths and characteristics of the individual models, thus enhancing the holistic understanding of the target task. Furthermore, the combination of diverse models can mitigate the risk of overfitting and enhance generalization capability.

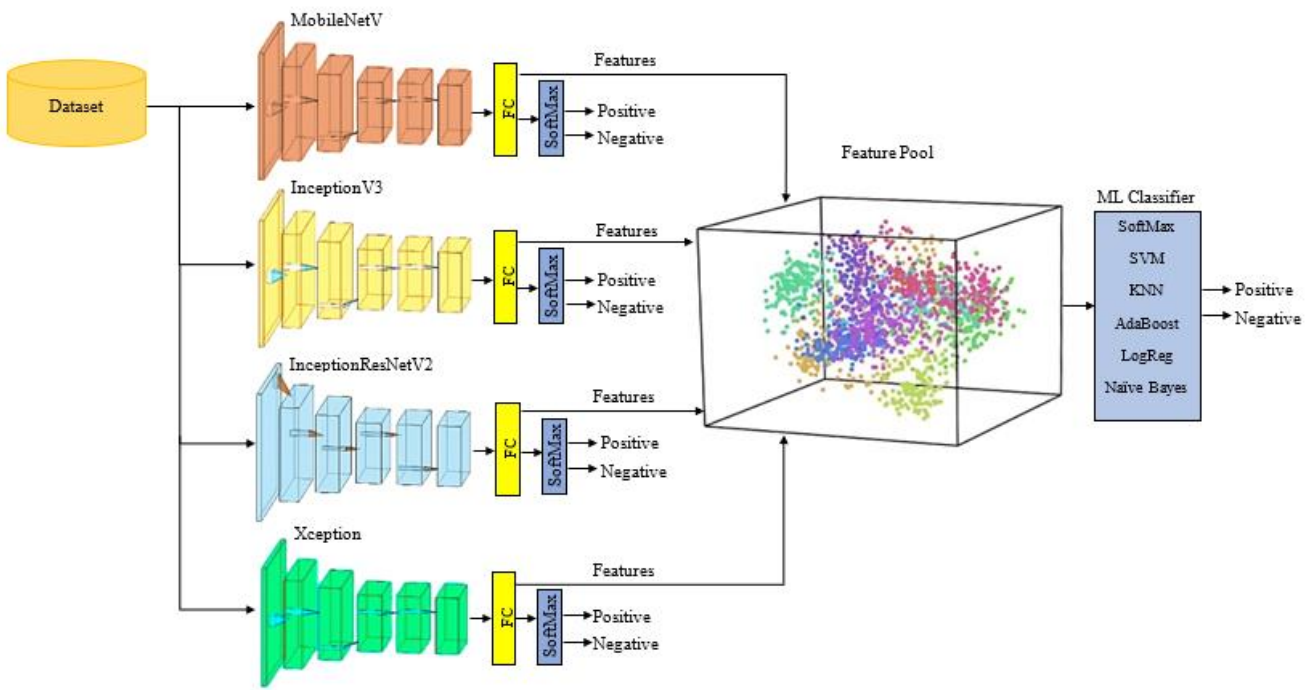

Figure 2. Block diagram of the proposed Deep Feature Fusion model

## 3.4 Part2: Proposed framework, Multi-task Classification

Traditional anomaly detection models struggle with handling new or unforeseen anomalies, making them susceptible to false negatives or misclassifications. In contrast to binary anomaly detection, which deals with only one class of anomalies and one class of normal data, multi-anomaly detection addresses scenarios where multiple types of anomalies are present.

The proposed multi-task classification approach introduces a critical innovation that eliminates the need to retrain the entire model from scratch when introducing a new anomaly class. This approach unifies features extracted from various CNN models and multiple datasets with different classes into a single feature space. These features are then used to train ML classifiers. Figure 3 provides a schematic diagram of the multi- task classification model. In the scenario presented in this paper, two feature fusion pools were used: the UCF-features pool and the RLVS-features pool, which fused the features extracted from the UCF and RLVS datasets. The UCF-feature pool consists of features related to normal and violent classes, while the RLVS-features pool includes features representing normal and shoplifting classes.

To categorize incoming frames as violent, shoplifting, or normal, the proposed approach utilized the UCF-features pool and RLVS-features pool, along with their corresponding class labels, to create a unified feature space. This feature space was then used to train the ML classifiers to categorize and classify incoming frames based on their respective classes.

There are several advantages to incorporating new classes without going through full retraining. This approach can save time and computational resources, especially when dealing with complex models or models that require significant training time. It also enables the system to adapt to emerging anomaly types, which enhances its robustness in dynamic environments. This approach effectively addresses a significant challenge in ML by allowing models to adapt to evolving anomaly patterns efficiently.

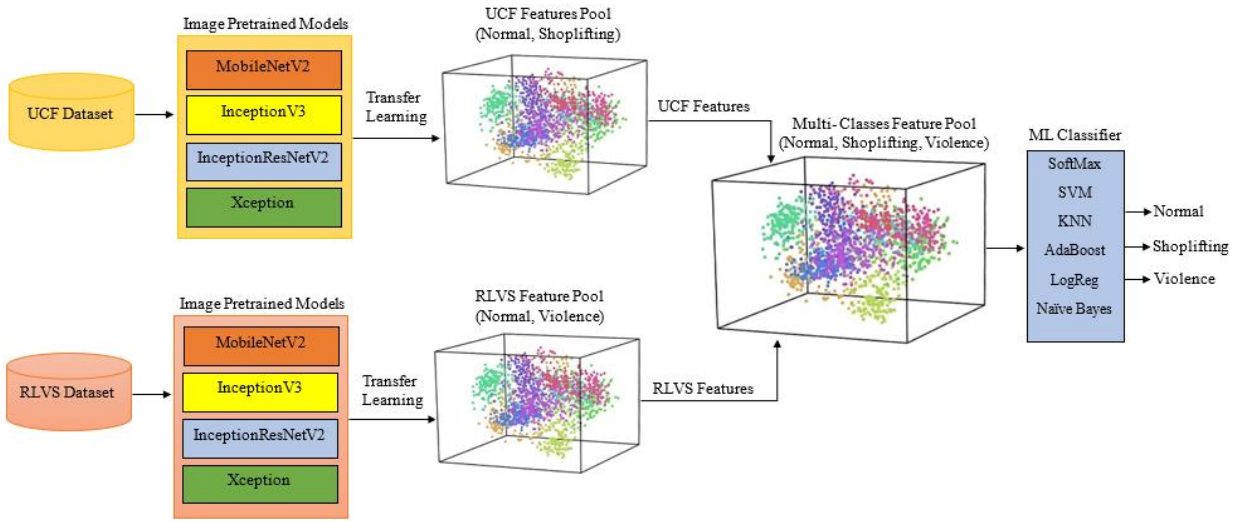

Figure 3. Schematic diagram of the proposed multi-classification model.

3.5 Training

We conducted experiments involving four individual CNN models: MobileNetV2, InceptionV3, InceptionResNetV2, and Xception, as well as a deep feature fusion model and a multi-task classification model. These experiments were structured as follows:

- Training and testing each individual CNN model on the UCF dataset.
- Training and testing each individual CNN model on the RLVS dataset.
- Assessing the performance of the deep fusion model through distinct tests on the UCF and RLVS datasets.
- Finally, evaluating the proposed multi-task classification approach combines the captured features from the UCF and RLVS datasets, each containing different abnormal behaviors.

## 3.6 Explainable tools

DL has been considered a complicated and obscure process, often described as a "black box" due to the difficulty in comprehending why a model makes specific choices. As a result, this lack of transparency can erode trust in the final decisions made by these systems [41]. Given this concern, this paper adopts the Grad-CAM and t-SNE visualization techniques to tackle these limitations and offer a more comprehensive insight into how DL methods arrive at their decisions.

1- The Grad-CAM (gradient-weighted class activation mapping) is an interpretability technique designed to elucidate the predictions of any DL model in a coherent and comprehensible manner. This is achieved through a focus on visualizing crucial regions within images. The approach capitalizes on gradients, effectively highlighting areas of the image that wield significant influence over the model's decision-making process. Grad-CAM starts with a forward pass, processing an image through a pre-trained CNN. Backpropagation calculates how changes in each feature map influence the final prediction score for a specific class. GAP computes importance scores for each feature map by averaging gradients. These important scores are used as weights for a weighted sum of feature maps, indicating their impact. The ReLU activation focuses on positive contributions, and the final output is a heatmap highlighting image regions that influenced CNN's decision. Brighter heatmap areas correspond to more influential image regions in classification [42][43]. Researchers can analyze and comprehend the crucial regions to better understand the model's reasoning process. This approach also helps them to verify whether the influential regions align with their expectations, boosting confidence in the model's predictions. Any spelling, grammar, or punctuation errors have been corrected.
2- t-SNE, which stands for t-Distributed Stochastic Neighbor Embedding, is a non-linear technique for reducing the dimensions of data while preserving the structure at various scales. It is particularly well-suited for visualizing high-dimensional datasets. The low-dimensional representation produced by t-SNE can be plotted, allowing you to visualize clusters, patterns, and relationships in the data that might be difficult to discern in the high-dimensional space. This paper uses t-SNE to understand the fusion techniques and how they improve the feature space.

## 4. Experimental Results

This section discusses the experimental results of various deep learning models used for anomaly detection, specifically focusing on tasks such as violence and shoplifting detection in video datasets.

It outlines the performance of individual CNN models and a proposed deep feature fusion model, multi-task classification approach assessed using metrics like accuracy, recall, precision, and F1-score.

## 4.1 Evaluation Metrics

The utilized models in this study underwent assessment employing a range of evaluation metrics, encompassing accuracy, recall, precision, and the F1-score. AD systems require minimized rates of false positives (FP) and false negatives (FN) while simultaneously maximizing the counts of true positives (TP) and true negatives (TN). TN signifies the count of correctly labeled negative (normal) instances, whereas TP corresponds to correctly labeled positive (anomaly) instances. Conversely, FP and FN counts reveal the number of instances inaccurately labeled as positive or negative [3]. Each evaluation metric is computed as follows:

$$\text{Accuracy} = \frac{TP + TN}{TP + TN + FP + FN} \quad (1)$$

$$\text{Recall} = \frac{TP}{TP + FN} \quad (2)$$

$$\text{Precision} = \frac{TP}{TP + FP} \quad (3)$$

$$\text{F1 score} = 2 \times \frac{\text{Precision} \times \text{Recall}}{\text{Precision} + \text{Recall}} \quad (4)$$

## 4.2 Experimental Results of Individual CNN models

The four deep CNN models used in this work were evaluated for performance in VAD tasks by being tested on the UCF and RLVS datasets, as described in subsequent sections.

### 4.2.1 Experiment Results on UCF Dataset

The results of the pre-trained CNN models were evaluated based on their accuracy, loss curves for training and validation, and the confusion matrix, as shown in Figure 4. Table 1 presents the evaluation metrics obtained by testing these models on the UCF dataset. MobileNetV2 performed the best in accuracy, precision, and F1 score, making it the ideal choice when minimizing false positives is a priority. Xception, on the other hand, had high recall but lower precision, making it a suitable option when capturing true positives is a priority. InceptionV3 and InceptionResNetV2 achieved a balance between these two metrics. Additionally, Figure 5 displays the Grad-CAM-generated heatmaps for these individual models.

Table 1. The experimental results of the individual models on the UCF dataset.

| Model | Accuracy (%) | Recall (%) | Precision (%) | F1 score (%) |
|---|---|---|---|---|
| MobileNet | 83.48 | 74.56 | 90.14 | 81.61 |
| Inception | 80.71 | 70.58 | 87.81 | 78.26 |
| InceptionResNet | 79.60 | 78.15 | 79.92 | 79.03 |
| Xception | 70.50 | 89.10 | 64.47 | 74.81 |

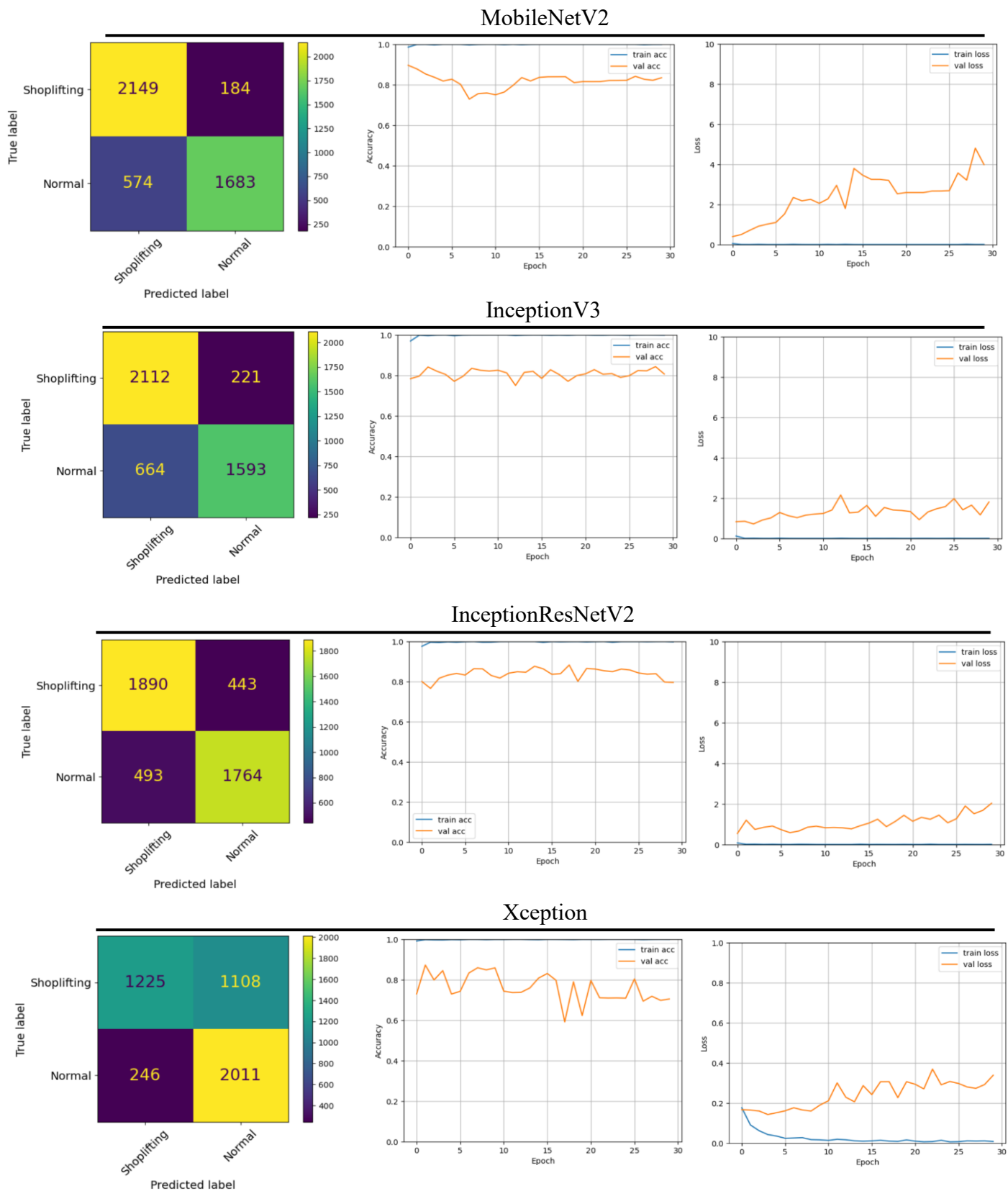

Figure 4. Confusion matrix, loss, and accuracy of the individual CNN models on the UCF dataset.

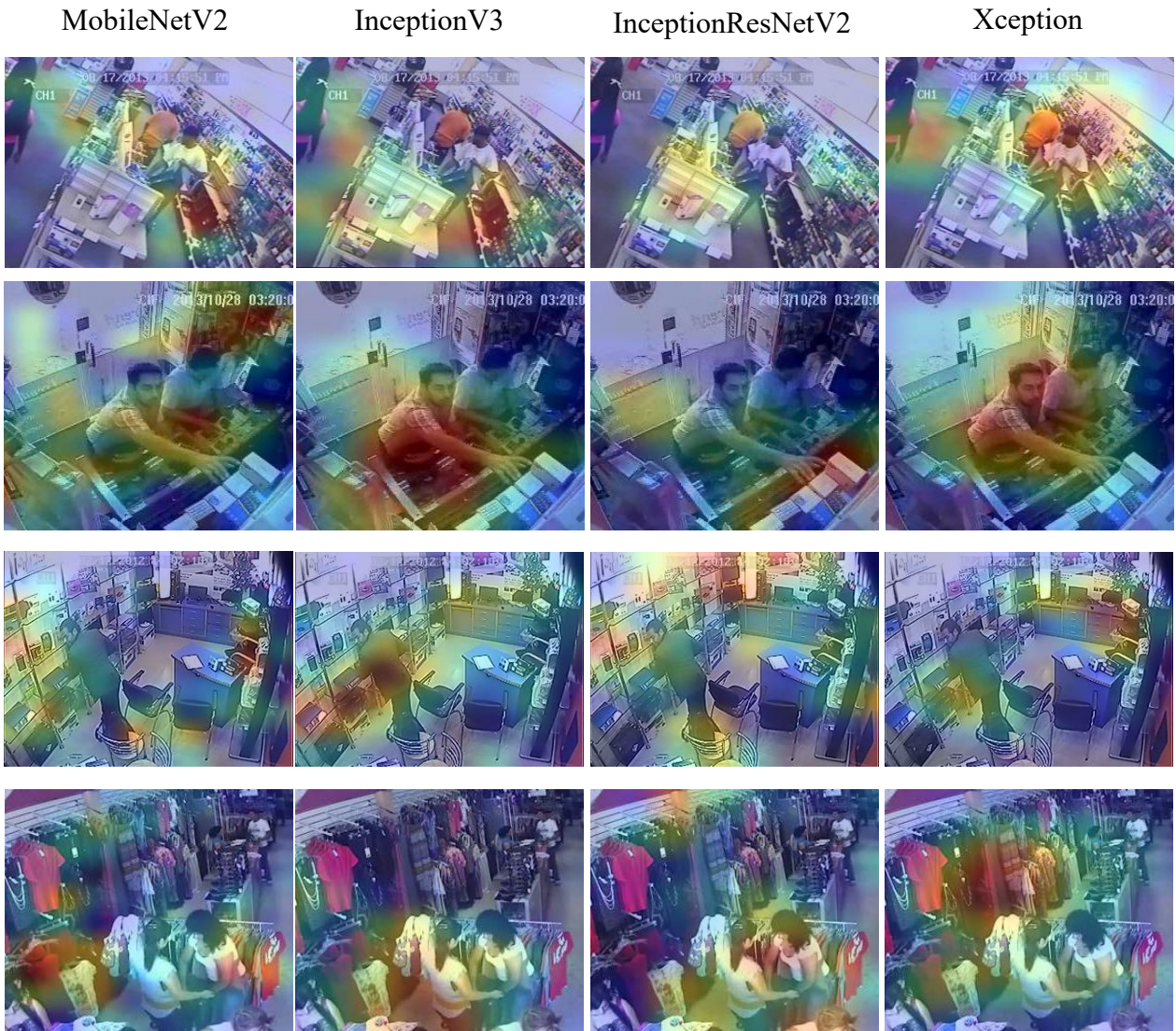

Figure 5. Grad-Cam with heatmap of shoplifting behaviour in the UCF dataset using the Individual models

### 4.2.2 Experiment Results on RLVS Dataset

The RLVS dataset served as the training and testing data for four distinct models in this specific scenario. These models underwent training and validation across multiple epochs, allowing the measurement of losses, accuracies, and confusion matrices to assess their performance. The results of this evaluation are presented in Figure 6. Table 2 provides a summary of the performance metrics of these models on the RLVS dataset. All four models demonstrated high accuracy, highlighting their proficiency in correctly classifying videos as either violent or normal. The differences in accuracy, recall, and F1 score between the models are minimal, with MobileNet holding a slight edge, demonstrating a strong capability to detect violent instances. InceptionResNet exhibited the highest precision. These models deliver robust performance in violence recognition, characterized by their high accuracy. Figure 7 displays the heatmap generated by Grad-CAM.

**Table 2.** The experimental results of the individual models on the RLVS dataset.

| Model | Accuracy (%) | Recall (%) | Precision (%) | F1 score (%) |
|---|---|---|---|---|
| MobileNet | 96.57 | 97.74 | 95.51 | 96.61 |
| Inception | 96.0 | 95.88 | 96.18 | 96.0 |
| InceptionResNet | 96.19 | 95.32 | 97.0 | 96.16 |
| Xception | 96.17 | 96.75 | 95.65 | 96.20 |

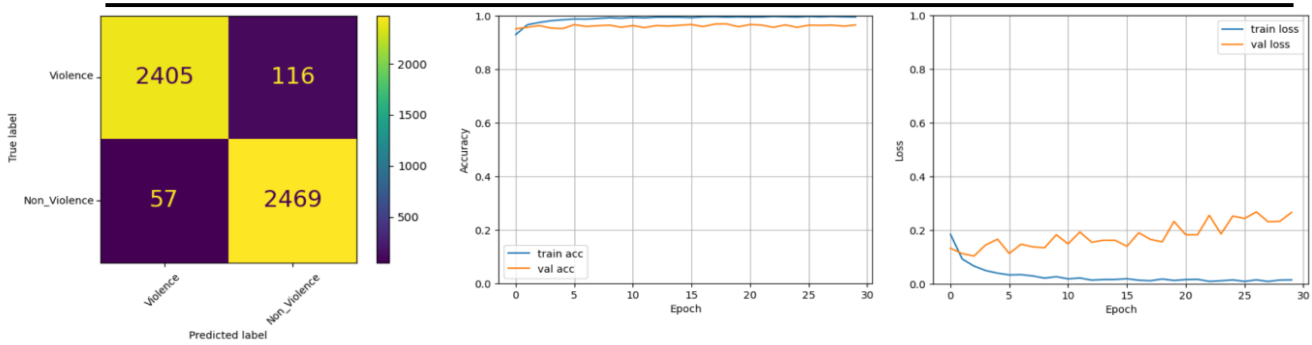

MobileNetV2

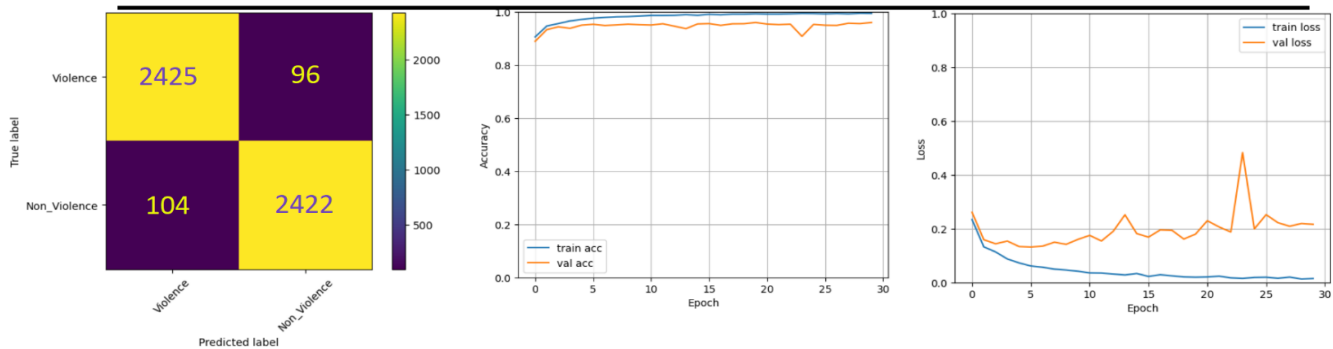

InceptionV3

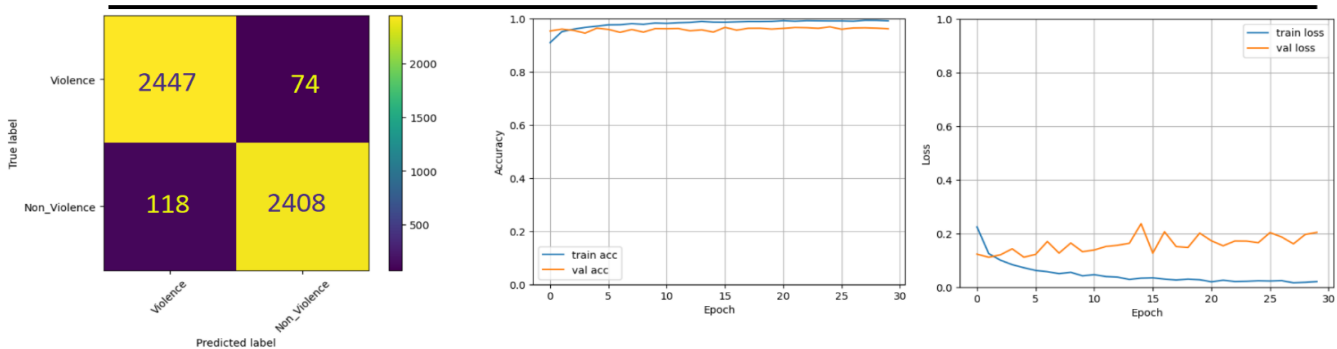

InceptionResNetV2

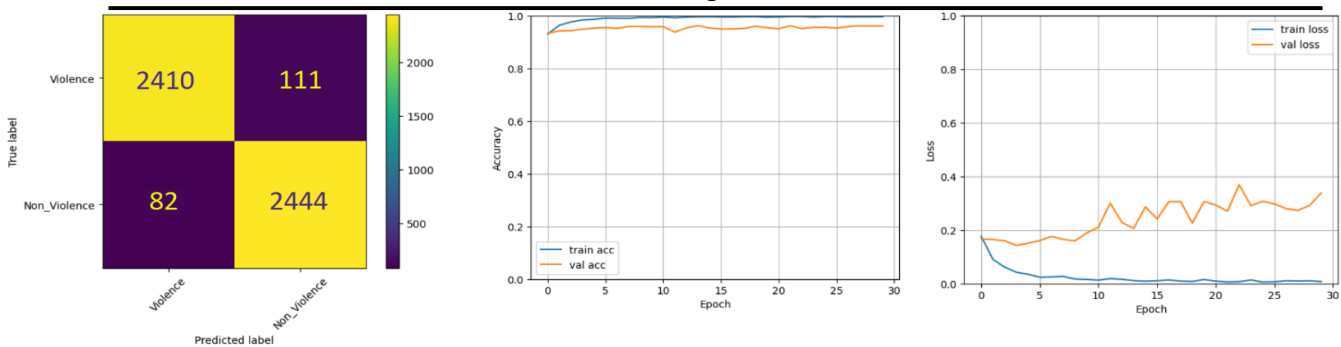

Xception

Figure 6. Confusion matrix, loss, and accuracy of the individual models on RLVS dataset.

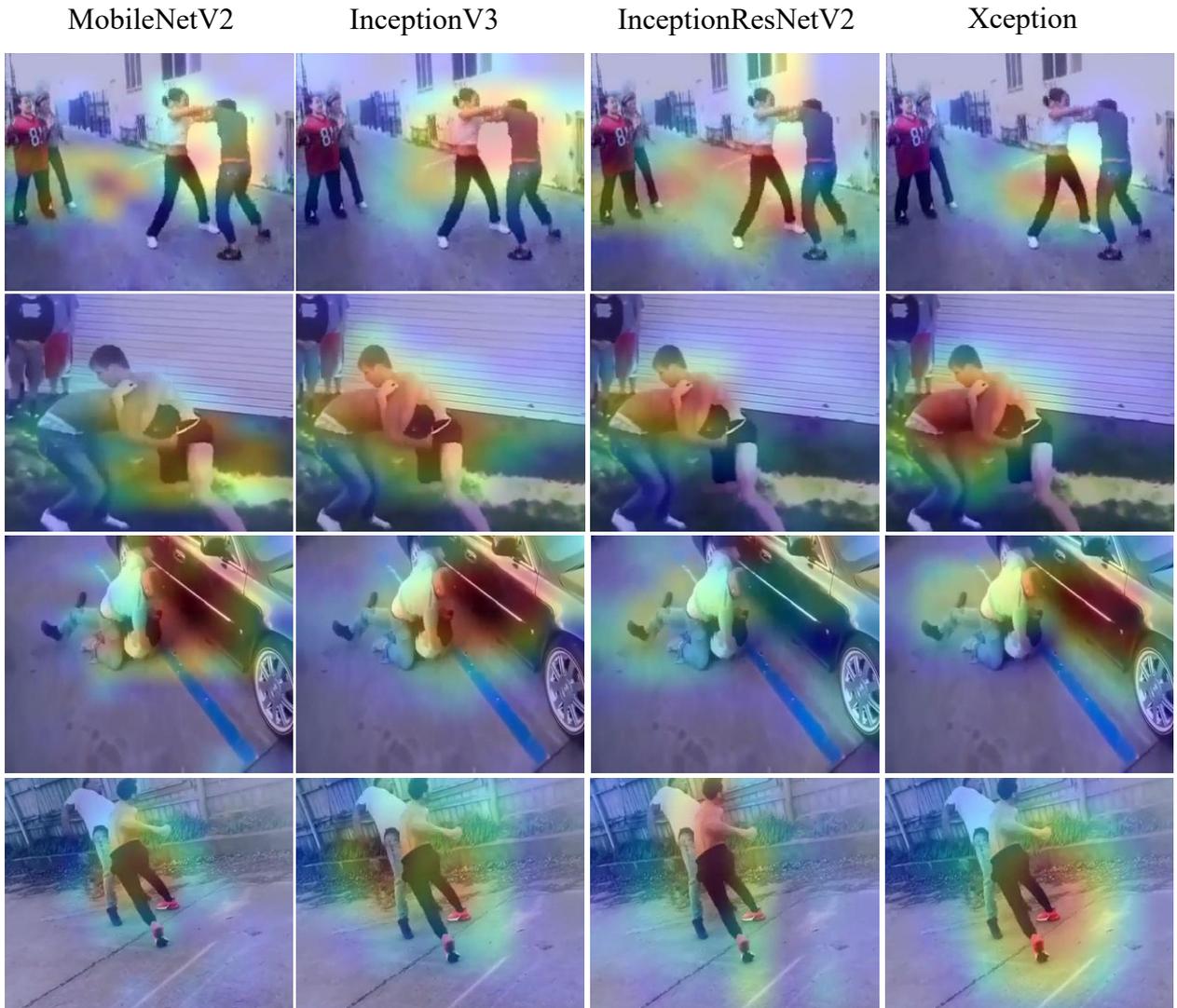

Figure 7. Grad-Cam with heatmap of violent behaviour in RLVS dataset using the Individual models

## 4.3. Experimental Results of the deep feature fusion Model

In the following subsections, we provide detailed information about the experimental results achieved by utilizing the deep Fusion model on both the UCF and RLVS datasets. Each model focused on a distinct region of interest, and the fusion of these four models proved to be highly effective in capturing features for the ML classifiers.

4.3.1 Experimental Results of the Deep Fusion Model on the UCF Dataset

The proposed feature fusion model was evaluated on the UCF dataset to evaluate its effectiveness in recognizing shoplifting behavior in surveillance videos. As mentioned earlier, this model involved combining features extracted from individual CNN models, which were trained on the UCF dataset, into a unified feature pool. Subsequently, we assessed the model's performance on the UCF testing dataset using six classifiers. Table 3 and Figure 8 present the results and their corresponding confusion matrices. The results demonstrate that the proposed model outperforms the individual CNN models, achieving an accuracy of 83.59%, a recall of 84.62%, a precision of 82.46%, and an F1 score of 83.53%. This represents an improvement of 0.11% over the MobileNetV2 model, which achieved the highest accuracy among the individual models. Figure 9 shows the Feature distribution visualized using t-SNE for UCF dataset including CNN models and features after fusion.

Table 3. Experimental results of the deep Fusion model on the UCF dataset

| Classifier | Accuracy (%) | Recall (%) | Precision (%) | F1 score (%) |
| --- | --- | --- | --- | --- |
| LogReg | 83.59 | 84.62 | 82.46 | 83.53 |
| SoftMax | 83.55 | 84.53 | 82.45 | 83.48 |
| KNN | 83.39 | 82.41 | 83.59 | 82.99 |
| SVM | 81.28 | 72.70 | 87.10 | 79.25 |
| AdaBoost | 81.28 | 72.70 | 87.10 | 79.25 |
| Naïve Bayes | 77.25 | 53.92 | 99.67 | 69.98 |

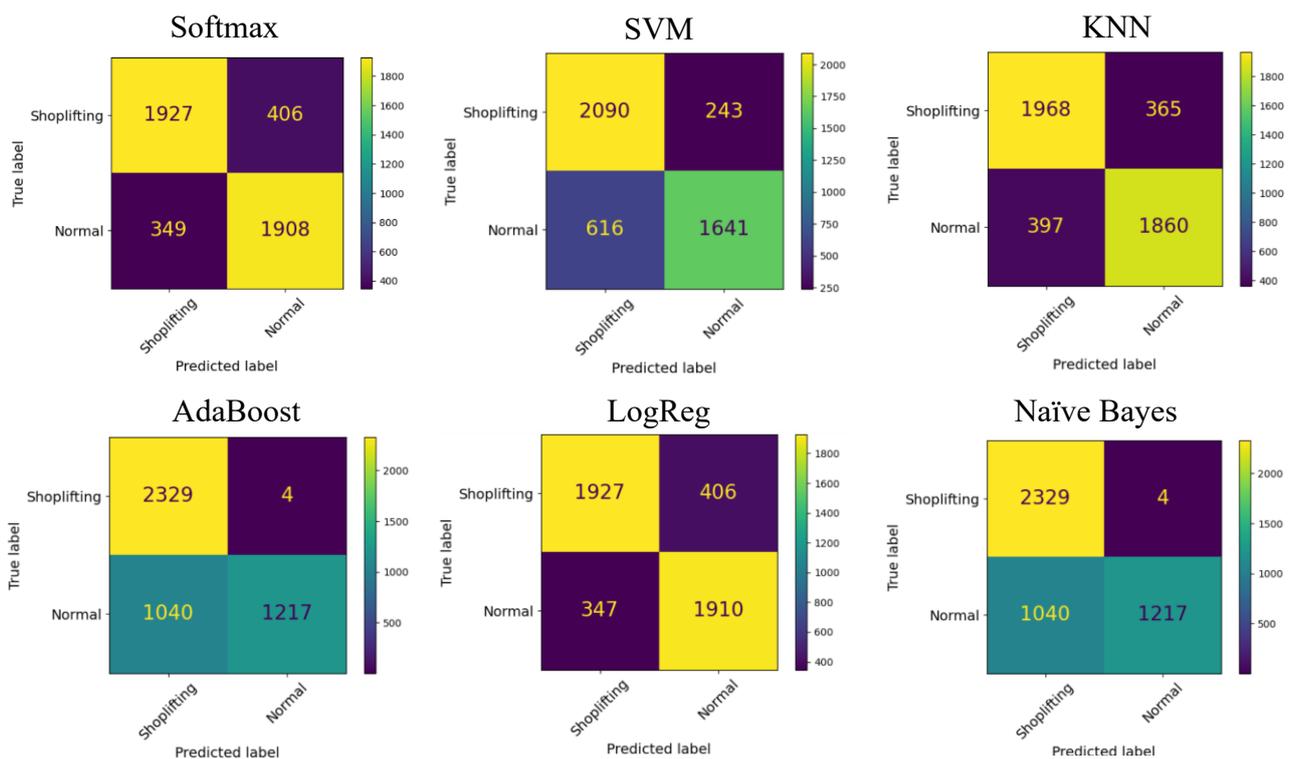

Figure 8. Confusion matrixes of ML classifiers on the UCF dataset.

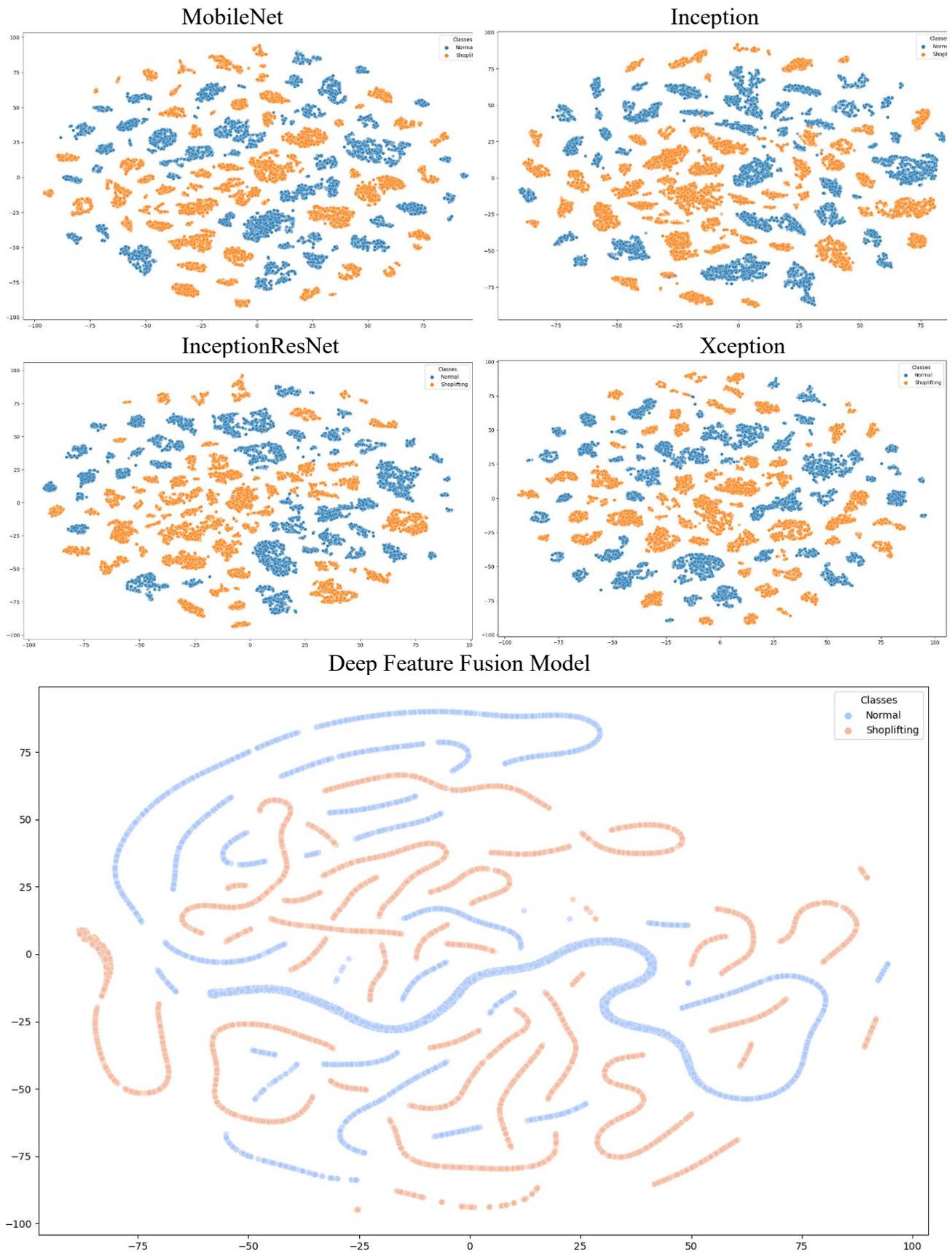

Figure 9: Feature distribution visualized using t-SNE for UCF dataset.

4.3.2 Experimental Results of The Deep Feature Fusion Model on the RLVS Dataset

Once again, the proposed deep feature fusion approach outperformed the individual CNN models used in this work when tested on the RLVS dataset for detecting violent activities in videos, achieving an accuracy of 97.99%. This represents an increase of 1.42% over the MobileNetV2 model, which achieved the highest accuracy among the individual models. Table 4 and Figure 10 present the experimental results, including the confusion matrices of the ML classifiers. Figure 11 shows the Feature distribution visualized using t-SNE for RLVS dataset including CNN models and features after fusion.

Table 4. Experimental results of the deep Fusion model on the RLVS dataset.

| Classifier | Accuracy (%) | Recall (%) | Precision (%) | F1 score (%) |
|---|---|---|---|---|
| KNN | 97.99 | 98.57 | 97.45 | 98.01 |
| LogReg | 97.89 | 98.57 | 97.26 | 97.91 |
| SoftMax | 97.89 | 98.57 | 97.26 | 97.91 |
| AdaBoost | 97.82 | 97.78 | 98.86 | 97.82 |
| SVM | 97.60 | 99.01 | 96.30 | 97.63 |
| Naïve Bayes | 97.34 | 99.28 | 95.57 | 97.39 |

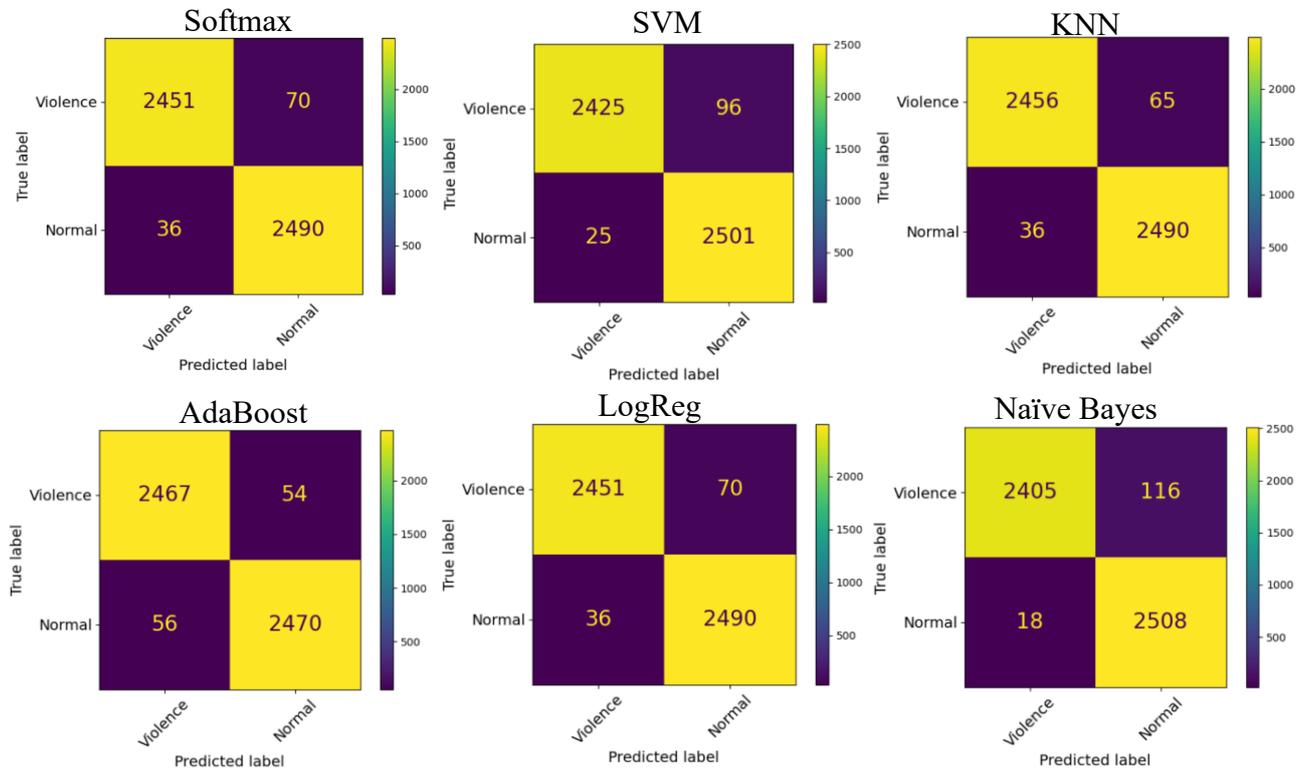

Figure 10. Confusion Matrices of ML Classifiers on the RLVS Dataset.

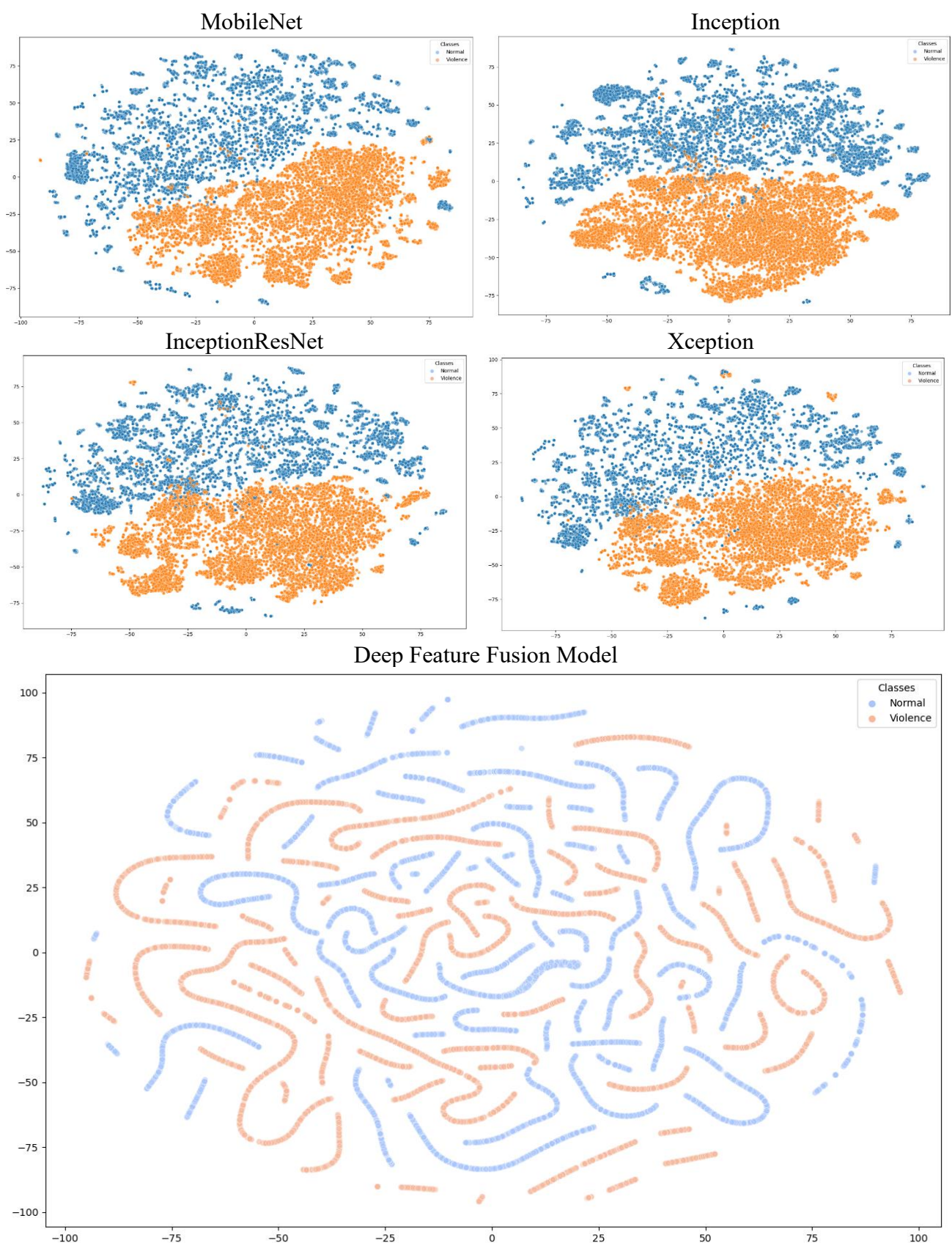

Figure11: Feature distribution visualized using t-SNE for RLVS dataset.

## 4.3. Experimental Results of the multi-task classification Model

We assessed the effectiveness of our multi-task classification model in identifying multiple anomaly classes by using two video anomaly behavior datasets - UCF, which has normal and shoplifting classes, and RLVS, which has normal and violent classes. The proposed model used four pre-trained CNN models as feature extractors to extract the features from the UCF and RLVS datasets. The extracted features from these different models for each dataset were fused to create a single feature set (Figure 12). It is worth noting that the features for normal behavior from both datasets were merged due to their similarity in behavior. After that, ML classifiers were trained to categorize and classify incoming frames as shoplifting, violent, or normal behavior. Table 5 shows our results, including accuracy, recall, precision, and F1 scores for six different classifiers - AdaBoost, KNN, LogReg, SoftMax, Naïve Bayes, and SVM. We also included their respective confusion matrices in Figure 13. AdaBoost showed the highest accuracy at 88.37%, recall at 84.34%, precision at 88.0%, and an F1 score of 85.43%, indicating strong overall performance. KNN also performed well with a high accuracy of 86.45%, recall of 82.05%, precision of 82.62%, and an F1 score of 82.08%. LogReg and SoftMax yielded similar performance metrics with moderate accuracy and F1 scores. SVM and Naïve Bayes showed lower accuracy, recall, precision, and F1 scores compared to the other classifiers.

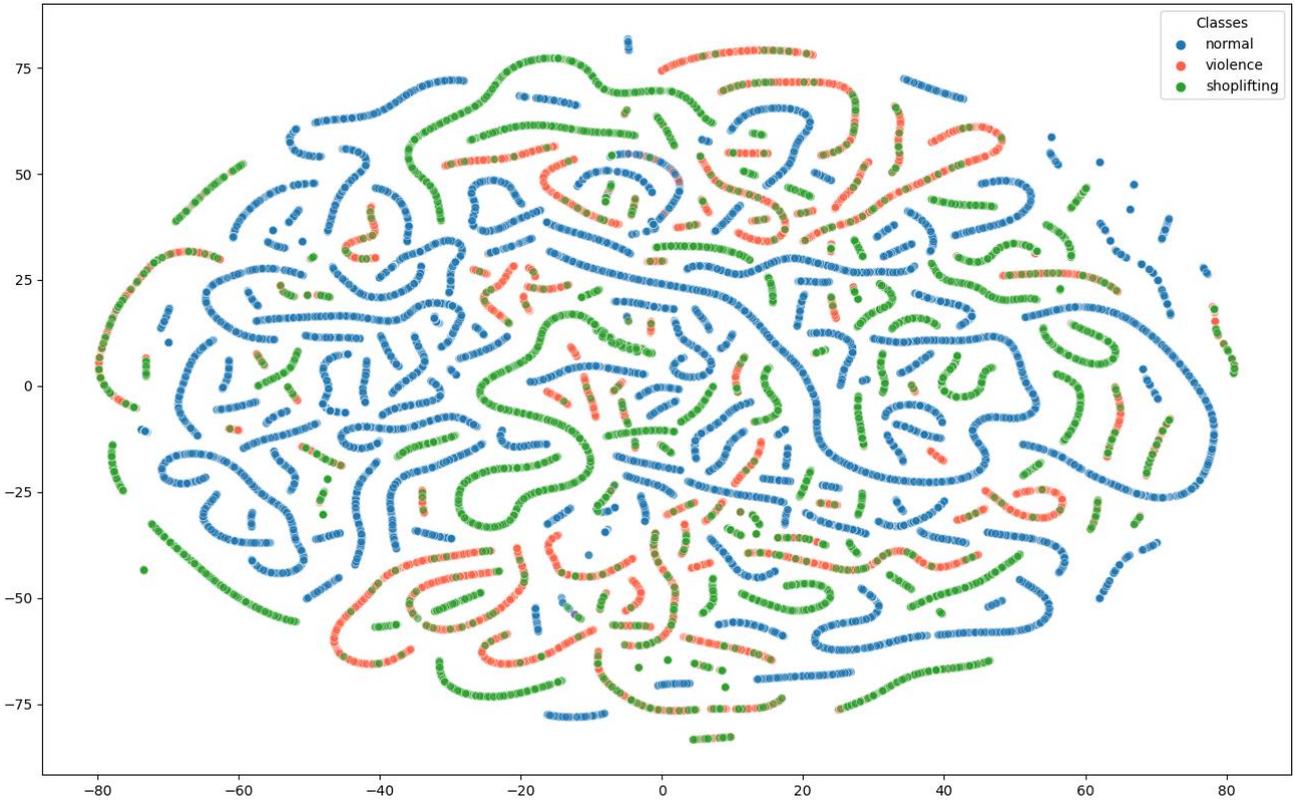

Figure 12: Feature distribution visualized using t-SNE for both datasets.

Table 5. Experimental results of the multi-classification model

| Classifier | Accuracy (%) | Recall (%) | Precision (%) | F1 score (%) |
| --- | --- | --- | --- | --- |
| AdaBoost | 88.37 | 84.34 | 88.00 | 85.43 |
| KNN | 86.45 | 82.05 | 82.62 | 82.08 |
| LogReg | 78.49 | 71.98 | 72.01 | 71.27 |
| SoftMax | 78.49 | 71.98 | 72.01 | 71.27 |
| SVM | 72.86 | 64.86 | 63.86 | 62.02 |
| Naïve Bayes | 71.87 | 63.70 | 62.17 | 60.35 |

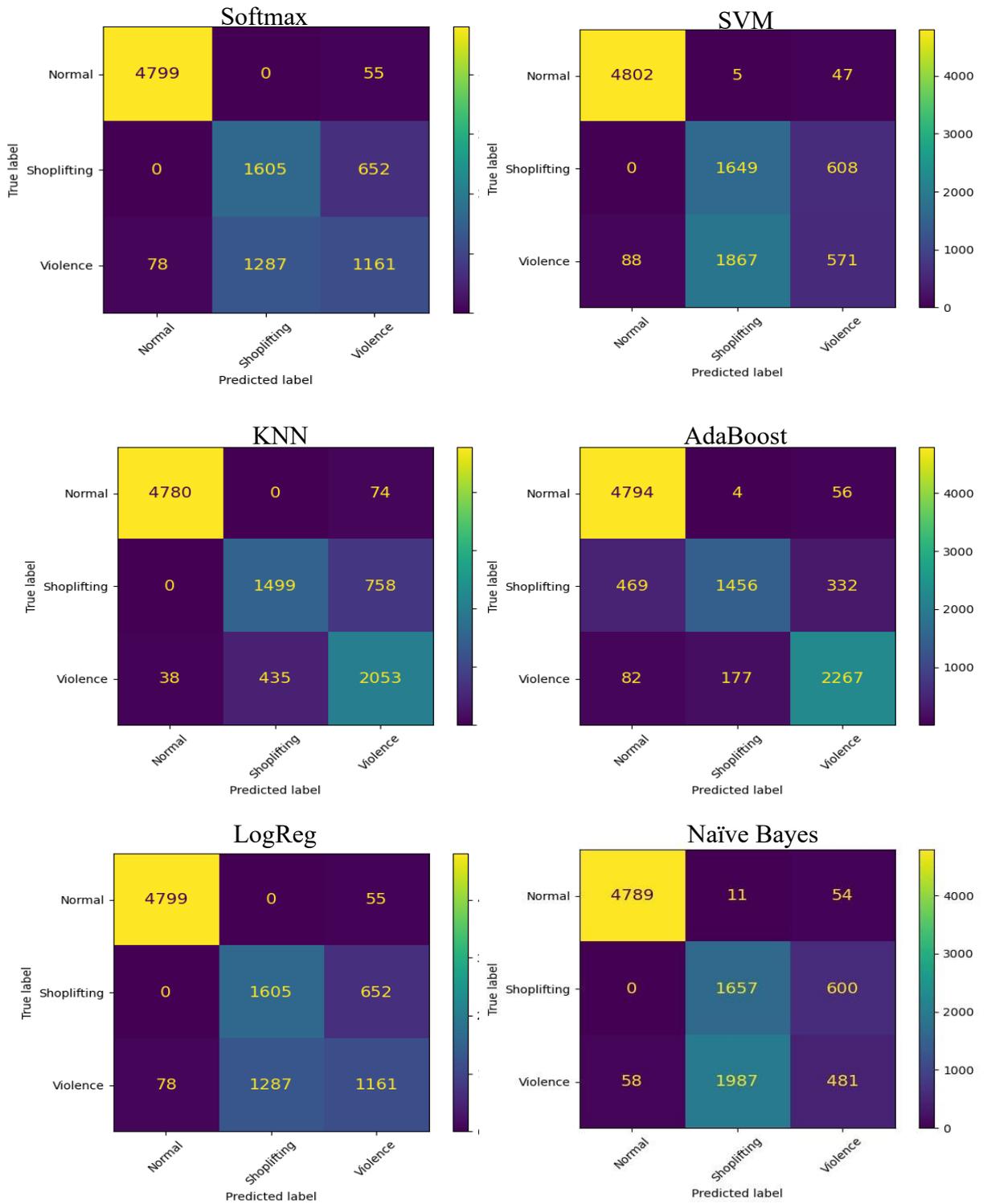

Figure 13. Confusion Matrices of ML Classifiers using multi-task classification Model.

## 4.4 Comparative Studies with Recent Research

In this section, we compared the performance of the proposed deep feature fusion approach to recent research that utilizes DL models to detect anomalous behaviors in surveillance videos, particularly those related to violence and shoplifting crimes. Table 6 presents a comparison of the accuracy values achieved by the proposed deep feature fusion model with existing methods for automatic shoplifting detection using the UCF dataset. On the other hand, table 7 presents the results of the experiments conducted by applying our deep feature fusion model and various DL methods for violence detection on the RLVS dataset.

Our proposed Deep Fusion approach has demonstrated remarkable superiority over contemporary methodologies by achieving notable accuracies of 83.59% and 97.99% on the UCF and RLVS datasets, respectively. These substantial performance metrics unequivocally affirm our approach's outstanding quality and efficacy in VAD. Such results establish our method's prowess and signify a substantial stride forward in advancing the current state-of-the-art in this intricate field.

The exceptional accuracy attained on both datasets underscores our approach's potential to address real-world challenges in video surveillance and anomaly detection scenarios, signifying its value in practical applications and its potential for further research and development.

Table 6. Comparison of the deep feature fusion model with other methods using the UCF dataset.

| Ref., year | Model | Accuracy % |
|---|---|---|
| [12], 2020 | 3D CNN | 75.0 |
| [16], 2021 | 3D CNN | 75.7 |
| [17], 2021 | InceptionV3 and LSTM networks | 74.53 |
| [28], 2023 | InceptionV3 and bidirectional LSTM | 81.0 |
| **Proposed deep feature fusion model** | | **83.59** |

Table 7. Comparison of the deep feature fusion model with other methods using the RLVS dataset.

| Ref., year | Model | Accuracy % |
|---|---|---|
| [36], 2019 | VGG16 + LSTM | 88.20 |
| [11], 2020 | ValdNet2 + GRU | 96.74 |
| [44], 2021 | Flow Gated RGB | 87.25 |
| [21], 2022 | keyframe + ResNet18 network | 94.60 |
| [22], 2022 | 3DCNN + LSTM networks | 96.50 |
| [26], 2023 | BoTNet152 + TCN | 93.15 |
| [33], 2024 | MultiWave-Net | 96.0 |
| [34], 2024 | MLP-Mixer architecture | 96.0 |
| **Proposed deep feature fusion model** | | **97.99** |

The model for multi-classification achieved a high accuracy of 88.37% by using the AdaBoost classifier. Its primary task was to recognize three classes of behavior, which included two distinct types of abnormal behavior (shoplifting and violence) and a normal class. The model leveraged the UCF and RLVS datasets, making it highly effective in accurately identifying anomalies. It provided a robust solution for multi-anomaly recognition and significantly enhanced the capacity to detect and classify diverse anomalous behaviors across various scenarios. This approach is groundbreaking and innovative as it addresses the challenge of integrating multi-tasks within video anomaly systems without requiring complete model retraining when incorporating a new anomaly class. As a result, our approach is unique and incomparable to prior studies regarding this specific aspect.

4.5 Independent Test

An independent test involves evaluating a model's performance on data it has not seen or been trained on, a crucial step in assessing the model's generalization capabilities. In this study, we conducted an independent test on the proposed multi-task classification model to evaluate its performance on unseen data with new scenarios and gauge its ability to generalize learned patterns to new information. The model underwent training on the UCF dataset, encompassing shoplifting actions, and the RLVS dataset, which includes instances of violent actions. The model was then tested on a Movie dataset [45] featuring both normal and violent actions. It's worth noting that, unfortunately, we could not acquire a dataset containing shoplifting actions for use in this particular test. Table 8 lists the results achieved by the multi-task classification demonstrated on the movie test set. KNN stands out with the highest accuracy at 87.25% among the presented classifiers. It also exhibits balanced recall at 80.25%, precision at 97%, and an F1 score at 87.87%, indicating robust performance across different metrics. AdaBoost also performs well across all metrics: accuracy at 86.23%, recall at 77.80%, precision at 99.43%, and F1 score at 87.30%. This noteworthy result suggests the model's proficiency in generalizing to sequences of violence across diverse scenarios, validating its robustness and potential practical utility. Figure 14 displays the heatmap generated by the Grad-CAM method.

Table 8. Experimental results of the multi-task classification model on the Movie dataset.

| Classifier | Accuracy (%) | Recall (%) | Precision (%) | F1 score (%) |
| --- | --- | --- | --- | --- |
| KNN | 87.25 | 80.25 | 97.00 | 87.87 |
| AdaBoost | 86.23 | 77.80 | 99.43 | 87.30 |
| SoftMax | 82.90 | 74.60 | 99.50 | 85.25 |
| LogReg | 82.90 | 74.60 | 99.50 | 85.25 |
| Naïve Bayes | 81.79 | 75.90 | 97.30 | 85.36 |
| SVM | 78.46 | 71.58 | 98.50 | 82.97 |

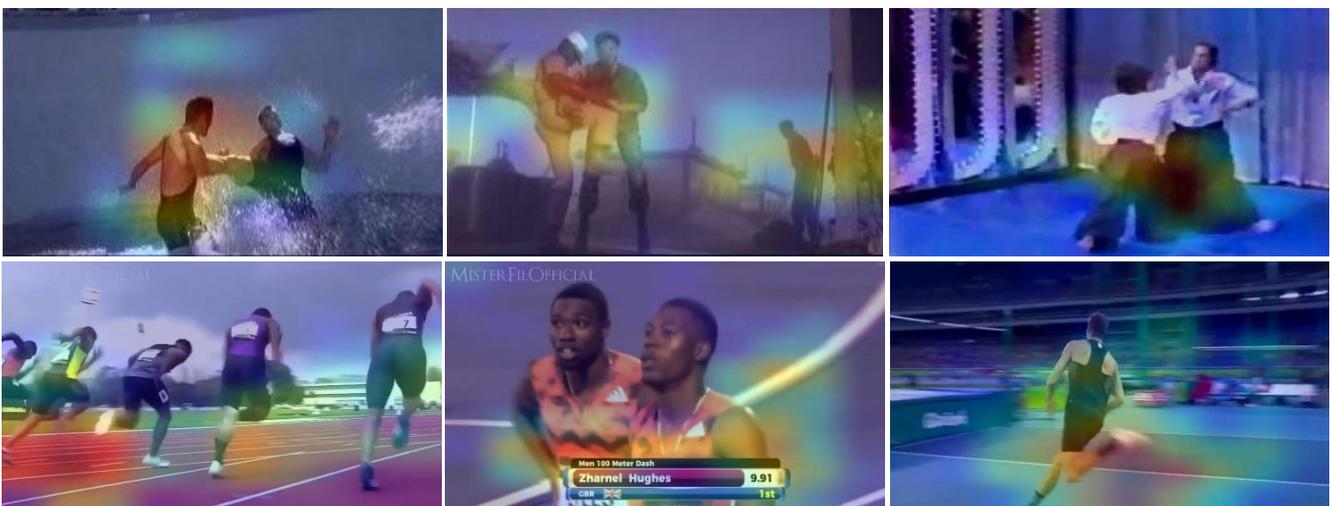

Figure 14. Grad-Cam with heatmap of the Movie dataset using the multi-task classification model. The first row presents violent behavior, and the second row presents normal behavior.

# 5. Conclusion

This paper addresses the challenging task of detecting anomalies in complex image data that is characterized by noise and diverse actions such as violence, shoplifting, and property destruction. While DL has shown promising results in this domain, previous studies have often needed help with the crucial problem of generalization across different AD tasks without resorting to training from scratch for each new task. This approach is not only time-consuming and computationally expensive but also unfair. To mitigate these issues, our paper introduces a novel DL framework that comprises three key components. First, TL is used to enhance feature generalization. Second, model fusion is employed to improve feature representation, which enhances generalization. Finally, multi-task classification is used to enable the generalization of the classifier across various tasks. Empirical results demonstrate the effectiveness of our approach, surpassing state-of-the-art methods. Using a single classifier, we achieved an impressive accuracy of 97.99% on the RLVS dataset for violence detection, 83.59% on the UCF dataset for shoplifting detection, and 88.37% on both datasets, all without the necessity of training from scratch for each task. To the best of our knowledge, this represents the first successful resolution of the generalization problem in anomaly detection, which is a significant advancement in this domain. In conclusion, our novel DL framework provides a more efficient and fair approach to AD in complex image data.